\documentclass{article}
    
\usepackage{PRIMEarxiv}

\usepackage[utf8]{inputenc} 
\usepackage[T1]{fontenc}    
\usepackage{hyperref}       
\usepackage{url}            
\usepackage{booktabs}       
\usepackage{amsfonts}       
\usepackage{nicefrac}       
\usepackage{microtype}      
\usepackage{fancyhdr}       
\usepackage{graphicx}       
\usepackage[square,numbers,comma,sort&compress]{natbib}
\usepackage{amsmath}
\usepackage{multirow}
\usepackage{tikz}

\usepackage{xcolor}

\title{Bielik-Minitron-7B: Compressing Large Language Models via Structured Pruning and Knowledge Distillation for the Polish Language}

\author{
  Remigiusz Kinas \\
  Bielik.AI, Ingenix.ai \\
  \texttt{remigiusz.kinas@bielik.ai} \\
  \And
  Paweł Kiszczak \\
  Bielik.AI, Vstorm \\
  \texttt{pawel.kiszczak@bielik.ai} \\
  \And
  Sergio P. Perez \\
  NVIDIA \\
  \texttt{sergiop@nvidia.com} \\
  \And
  Krzysztof Ociepa \\
  Bielik.AI, Azurro.pl \\
  \texttt{krzysztof.ociepa@bielik.ai} \\
  \And
  \L{}ukasz Flis \\
  Bielik.AI, ACK Cyfronet AGH \\
  \texttt{lukasz.flis@cyfronet.pl} \\
  \And
  Krzysztof Wr\'obel \\
  Bielik.AI, Jagiellonian University \\
  \texttt{krzysztof.wrobel@bielik.ai} \\
  \And
  Adrian Gwo\'zdziej \\
  Bielik.AI, ACK Cyfronet AGH \\
  \texttt{adrian.gwozdziej@bielik.ai} \\
}

\begin{document}
\maketitle

\begin{abstract}
This report details the creation of Bielik-Minitron-7B, a compressed 7.35B parameter version of the Bielik-11B-v3.0 model, specifically optimized for European languages. By leveraging a two-stage compression methodology inspired by the NVIDIA Minitron approach, we combined structured hybrid pruning and knowledge distillation to reduce the model's parameter count by 33.4\%, from 11.04B to 7.35B. We utilized the NVIDIA Model Optimizer for structural pruning and the NVIDIA NeMo Framework for logit-based distillation for quality recovery. Following distillation, the model underwent a rigorous alignment pipeline consisting of Supervised Fine-Tuning (SFT), Direct Preference Optimization (DPO-P), and Reinforcement Learning (GRPO). Our final model successfully recovered approximately 90\% of the baseline model's performance while providing up to 50\% inference speedup. This approach demonstrates an efficient pathway to create language models for less-represented languages, preserving the original model quality while reducing inference deployment costs.
\end{abstract}


\section{Introduction}
The progress on Large Language Models (LLMs) continues to transform natural language processing, leading to wider adoption and better model capabilities. As the size of the models grows, it is crucial to address the significant increase in computational resources to deploy them, especially in terms of GPU-accessible VRAM. For the European languages market and ecosystem, there is a critical need for models that maintain a balance between high-performance reasoning and deployment efficiency. 

An emerging technique to address the increasing size of LLMs is model compression. It aims to shrink the LLM so that it requires less memory and compute to deploy it, while preserving most of its original quality and behaviour. Today model compression has become an indispensable step when releasing a model, see, for instance, Qwen3 \cite{yang2025qwen3}, Nemotron-Nano-9B-v2 \cite{basant2025nvidia} or DeepSeek-R1 \cite{guo2025deepseek}. In addition, model compression is also employed to reduce the cost to post-train an open source model \cite{bercovich2025llama}: by reducing the size before the post-training, less compute is needed for stages such as supervised fine-tuning and reinforcement learning.

In collaboration with NVIDIA, we developed Bielik-Minitron-7B by compressing our flagship LLM Bielik-11B-v3.0 \cite{Bielik11Bv3i}. By enhancing our usual training pipeline, created and developed throughout the different iterations of the Bielik project, with structured pruning and distillation strategies rather than training a smaller-sized model from scratch, we significantly reduced the carbon footprint and computational cost of development while maintaining high linguistic fidelity, optimal size and quality of the final product.

\section{Related Work}

Our methodology is rooted in the NVIDIA Minitron approach \cite{sreenivas2024llm, muralidharan2024compact}, which demonstrates that structured pruning followed by distillation is superior to training smaller models from scratch. This section provides a comprehensive overview of neural network pruning approaches for LLMs, situating our work within the broader landscape of model compression techniques.

\subsection{Taxonomy of Pruning Methods}

Neural network pruning methods can be broadly categorized along two primary axes: \emph{pruning granularity} and \emph{importance estimation strategy}. In terms of granularity, methods fall into two major categories:

\begin{itemize}
    \item \textbf{Unstructured Pruning:} Removes individual weights regardless of their position within weight matrices. While achieving high compression ratios, unstructured sparsity typically requires specialized hardware or software (e.g., sparse matrix libraries) to realize actual speedups during inference.
    \item \textbf{Structured Pruning:} Removes entire architectural components such as attention heads, neurons, or transformer layers. This approach yields immediate hardware-friendly acceleration without requiring sparse computation support.
\end{itemize}

We now survey the most prominent pruning methods for LLMs, comparing their importance estimation strategies and practical applicability.

\subsection{Magnitude-Based Pruning}

The foundational approach to neural network pruning, introduced by Han et al.~\cite{han2015learning}, operates on a simple principle: weights with small absolute magnitudes contribute less to the network's output and can be safely removed. Given a weight matrix $\mathbf{W}$, the importance score for each weight $w_{ij}$ is computed as:
\begin{equation}
    S_{ij}^{\text{mag}} = |w_{ij}|
\end{equation}
Weights below a threshold $\tau$ are pruned: $w_{ij} = 0$ if $|w_{ij}| < \tau$. The method follows a three-stage pipeline: (1) train the network to convergence, (2) prune weights with smallest magnitudes, and (3) retrain to recover accuracy. Han et al. demonstrated 9$\times$ compression on AlexNet and 13$\times$ on VGG-16 without accuracy loss. However, for modern LLMs, the retraining phase becomes prohibitively expensive, motivating the development of one-shot pruning methods.

\subsection{SparseGPT: One-Shot Pruning via Optimal Brain Compression}

SparseGPT \cite{frantar2023sparsegpt} addresses the scalability challenge by enabling pruning of massive models (100B+ parameters) without any retraining. The method reformulates pruning as a sparse regression problem, solving for optimal weight updates that minimize the reconstruction error after removing weights.

For a layer with weight matrix $\mathbf{W}$ and calibration inputs $\mathbf{X}$, SparseGPT minimizes:
\begin{equation}
    \min_{\hat{\mathbf{W}}} \|\mathbf{W}\mathbf{X} - \hat{\mathbf{W}}\mathbf{X}\|_2^2 \quad \text{subject to sparsity constraint on } \hat{\mathbf{W}}
\end{equation}
The algorithm proceeds row-by-row through the weight matrix, using the Hessian $\mathbf{H} = \mathbf{X}\mathbf{X}^T$ to compute optimal weight updates that compensate for pruned weights. For each row, weights are greedily pruned while remaining weights are adjusted according to:
\begin{equation}
    \delta_{\text{row}} = -\frac{w_p}{[\mathbf{H}^{-1}]_{pp}} \cdot \mathbf{H}^{-1}_{:,p}
\end{equation}
where $w_p$ is the pruned weight and $[\mathbf{H}^{-1}]_{pp}$ is the corresponding diagonal element of the inverse Hessian. This enables 50-60\% unstructured sparsity on OPT-175B and BLOOM-176B in under 4.5 hours with negligible perplexity increase. SparseGPT also supports semi-structured patterns (2:4, 4:8) compatible with hardware acceleration.

\subsection{Wanda: Pruning by Weights and Activations}

Wanda (Pruning by \textbf{W}eights \textbf{and} \textbf{A}ctivations) \cite{sun2024wanda} provides an even simpler approach that requires no weight updates. The key insight is that a weight's importance depends not only on its magnitude but also on the magnitude of its corresponding input activations. The importance score combines both factors:
\begin{equation}
    S_{ij}^{\text{wanda}} = |w_{ij}| \cdot \|\mathbf{x}_j\|_2
\end{equation}
where $w_{ij}$ is the weight connecting input $j$ to output $i$, and $\|\mathbf{x}_j\|_2$ is the $\ell_2$-norm of the $j$-th input feature computed over a small calibration set. Pruning is performed \emph{per-output}, meaning for each output neuron $i$, the weights with smallest $S_{ij}^{\text{wanda}}$ scores are removed independently.

This formulation is motivated by the observation that LLMs exhibit emergent large-magnitude features---certain activation dimensions consistently show much larger values than others. By incorporating activation norms, Wanda naturally preserves weights connected to these important features. The method significantly outperforms magnitude pruning and performs competitively with SparseGPT while being simpler and faster (no Hessian computation required).

\subsection{ShortGPT: Depth Pruning via Block Influence}

ShortGPT \cite{men2024shortgpt} takes a fundamentally different approach by removing entire transformer layers rather than individual weights. The method introduces the \textbf{Block Influence (BI)} metric to quantify each layer's contribution to the model's representational transformation.

For layer $i$ with input hidden states $\mathbf{H}_i$ and output hidden states $\mathbf{H}_{i+1}$, the Block Influence score measures how much the layer transforms its input:
\begin{equation}
    \text{BI}_i = 1 - \frac{1}{N}\sum_{n=1}^{N} \text{cos\_sim}(\mathbf{H}_i^{(n)}, \mathbf{H}_{i+1}^{(n)})
    \label{eq:shortgpt}
\end{equation}
where $N$ is the number of calibration samples and $\text{cos\_sim}(\cdot,\cdot)$ denotes cosine similarity averaged over the sequence dimension. Layers with low BI scores (high similarity between input and output) are considered redundant and can be removed.

The pruning algorithm proceeds as follows: (1) compute BI scores for all layers using a calibration dataset, (2) rank layers by BI score in ascending order, and (3) remove the $k$ lowest-scoring layers. Empirically, ShortGPT found that layers in the middle-to-later portions of the network (e.g., layers 21-29 in LLaMA-2-7B) exhibit the highest redundancy. The method achieves state-of-the-art results for depth pruning and is orthogonal to quantization, enabling combined compression strategies.

\subsection{LLM-Pruner: Task-Agnostic Structural Pruning}

LLM-Pruner \cite{ma2024llmpruner} introduces a comprehensive framework for structured pruning that preserves multi-task capabilities. The method operates through three stages:

\textbf{Stage 1 - Dependency Discovery:} LLM-Pruner constructs dependency graphs to identify \emph{coupled structures}---groups of parameters that must be pruned together to maintain architectural validity. For example, in multi-head attention, the output projections of all heads feeding into the same residual connection form a coupled group.

\textbf{Stage 2 - Importance Estimation:} Group importance is computed using first-order Taylor expansion:
\begin{equation}
    I_g = \left| \sum_{w \in g} w \cdot \frac{\partial \mathcal{L}}{\partial w} \right|
\end{equation}
where $g$ denotes a coupled group and $\mathcal{L}$ is the loss on calibration data. This approximates the change in loss when removing the entire group. Alternatively, LLM-Pruner supports $\ell_1/\ell_2$ norm-based importance and Hessian-weighted variants.

\textbf{Stage 3 - Recovery:} A lightweight post-training phase using LoRA \cite{hu2022lora} on approximately 50K samples recovers performance. The entire pipeline (3 minutes pruning + 3 hours recovery) is significantly more efficient than retraining from scratch.

LLM-Pruner supports multiple granularities: block-wise (entire attention/MLP blocks), channel-wise (hidden dimensions), and layer-wise (transformer layers), making it a versatile framework applicable to LLaMA, BLOOM, and other architectures.

\subsection{Sheared LLaMA: Targeted Structured Pruning}

Sheared LLaMA \cite{xia2024sheared} introduces a principled approach to structured pruning that produces models matching specific target architectures. The method combines two key innovations:

\textbf{Targeted Structured Pruning:} Rather than heuristically selecting components to prune, Sheared LLaMA learns binary masks through $\ell_0$ regularization:
\begin{equation}
    \mathcal{L}_{\text{total}} = \mathcal{L}_{\text{LM}} + \lambda \cdot \|\mathbf{z}\|_0
\end{equation}
where $\mathbf{z}$ represents learnable mask variables for each prunable component (attention heads, hidden dimensions, intermediate dimensions, layers). The masks are relaxed to continuous values during training using the hard-concrete distribution, and components with near-zero masks are removed. Lagrangian relaxation dynamically adjusts $\lambda$ to meet target architecture specifications.

\textbf{Dynamic Batch Loading:} To accelerate convergence, Sheared LLaMA dynamically adjusts the sampling proportions across training domains (e.g., Wikipedia, code, books) based on per-domain loss. Domains where the model struggles receive more samples, improving data efficiency.

Starting from LLaMA-2-7B, Sheared LLaMA produces 1.3B and 2.7B parameter models that outperform comparably-sized models trained from scratch, while requiring only 3\% of the original pre-training compute.

\subsection{The Minitron Approach}

The Minitron methodology \cite{sreenivas2024llm, muralidharan2024compact}, which inspires the model compression strategy of this work, combines structured pruning across multiple axes with knowledge distillation for performance recovery. We describe the approach in detail as it directly informs our compression of Bielik-11B-v3.

\subsubsection{Multi-Axis Pruning}

Minitron prunes along four orthogonal dimensions:

\begin{enumerate}
    \item \textbf{Depth Pruning:} Removes entire transformer layers, reducing sequential computation.
    \item \textbf{Width Pruning (Hidden Dimension):} Reduces the embedding/hidden dimension $d_{\text{model}}$, affecting all linear projections.
    \item \textbf{Attention Pruning:} Reduces the number of attention heads $n_{\text{heads}}$ or key-value heads in grouped-query attention.
    \item \textbf{MLP Pruning:} Reduces the intermediate dimension $d_{\text{ff}}$ of feed-forward layers.
\end{enumerate}

\subsubsection{Activation-Based Importance Estimation}

Component importance in Minitron is estimated using purely activation-based metrics that require only forward passes over a calibration dataset---no gradient computation is needed. For each prunable component (attention heads, MLP neurons, embedding channels), the framework collects activation magnitudes and computes importance scores by aggregating statistics across calibration samples. Specifically, the importance of a component is determined by the magnitude of its activations: components with consistently low activation magnitudes are considered least critical and are therefore candidates for removal. The concrete formulas for hidden dimension and FFN intermediate dimension importance are detailed in Section~\ref{sec:neuron-selection}.

\subsubsection{Neuron Selection Mechanism}
\label{sec:neuron-selection}

For width pruning, Minitron computes per-neuron importance scores by aggregating activation statistics during forward passes:

\textbf{Hidden Dimension Importance:} Hooks attached to layer normalization outputs collect activation magnitudes:
\begin{equation}
    I_j^{\text{hidden}} = \sum_{\ell=1}^{L} \left( \frac{1}{N} \sum_{n=1}^{N} |a_{\ell,j}^{(n)}| \right)^2
\end{equation}
where $a_{\ell,j}^{(n)}$ is the activation of neuron $j$ at layer $\ell$ for sample $n$. Neurons with highest importance scores are retained.

\textbf{FFN Intermediate Dimension Importance:} Similarly, for MLP layers:
\begin{equation}
    I_j^{\text{ffn}} = \sum_{\ell=1}^{L} \left( \frac{1}{N} \sum_{n=1}^{N} |h_{\ell,j}^{(n)}| \right)^2
\end{equation}
where $h_{\ell,j}^{(n)}$ is the intermediate activation at position $j$.

The top-$k$ neurons according to these importance scores are selected, and weight matrices are sliced accordingly using \texttt{torch.index\_select} operations.

\subsubsection{Depth Pruning via Block Influence}

For layer removal, Minitron employs a Block Influence strategy similar to ShortGPT, see equation~\eqref{eq:shortgpt}.
Layers with lowest BI scores (indicating minimal transformation of hidden states) are candidates for removal. Alternatively, an iterative approach evaluates different pruning configurations and selects the one with minimal performance degradation on a validation task.

\subsubsection{Knowledge Distillation Recovery}

After pruning, the student model is initialized with the surviving weights from the teacher and undergoes knowledge distillation. Following the empirical findings of \cite{muralidharan2024compact}, the training objective uses a \emph{logit-only} forward KL divergence loss, ignoring both the standard cross-entropy loss against ground-truth labels and intermediate state matching:
\begin{equation}
    \mathcal{L} = \text{KL}\left(\sigma(\mathbf{z}_t/T) \| \sigma(\mathbf{z}_s/T)\right)
\end{equation}
where $\mathbf{z}_t$ and $\mathbf{z}_s$ are the teacher and student logits respectively, $\sigma$ denotes the softmax function, and $T$ is a temperature hyperparameter that softens the output distributions. This formulation enables the student to learn directly from the teacher's full probability distribution over the vocabulary, capturing inter-token relationships and confidence calibration that would be lost with hard-label training alone. The authors of \cite{muralidharan2024compact} demonstrate that logit-only distillation is sufficient for high-quality recovery when depth reduction is moderate, making it well-suited for our configuration.

Crucially, Minitron achieves strong results using less than 3\% of the original pre-training data, making it highly efficient compared to training from scratch. The resulting models achieve 2-4$\times$ compression with up to 16\% improvement in MMLU compared to equivalently-sized models trained de novo.

\subsection{Summary of Pruning Approaches}

Table~\ref{tab:pruning_comparison} summarizes the key characteristics of the surveyed pruning methods.

\begin{table*}[ht]
    \centering
    \begin{tabular}{lccccc}
        \toprule
        \textbf{Method} & \textbf{Granularity} & \textbf{Importance Metric} & \textbf{Retraining} & \textbf{One-Shot} & \textbf{Hardware-Friendly} \\
        \midrule
        Magnitude \cite{han2015learning} & Unstructured & $|w|$ & Required & No & No \\
        SparseGPT \cite{frantar2023sparsegpt} & Unstructured/Semi & Hessian-based & No & Yes & Partial (2:4) \\
        Wanda \cite{sun2024wanda} & Unstructured/Semi & $|w| \cdot \|x\|$ & No & Yes & Partial (2:4) \\
        ShortGPT \cite{men2024shortgpt} & Structured (Depth) & Block Influence & Optional & Yes & Yes \\
        LLM-Pruner \cite{ma2024llmpruner} & Structured & Taylor / $\ell_1$/$\ell_2$ & Light (LoRA) & Yes & Yes \\
        Sheared LLaMA \cite{xia2024sheared} & Structured & Learned $\ell_0$ masks & Integrated & No & Yes \\
        Minitron \cite{muralidharan2024compact} & Structured (Hybrid) & Activation + BI & Distillation & Yes & Yes \\
        \bottomrule
    \end{tabular}
    \caption{Comparison of LLM pruning methods.}
    \label{tab:pruning_comparison}
\end{table*}

\subsection{Rationale for Selecting Minitron}

Among the surveyed pruning methodologies, we selected the \textbf{Minitron approach} for compressing Bielik-11B-v3.0 based on several strategic and technical considerations. First, Minitron offers a \emph{hybrid pruning} framework that simultaneously optimizes across multiple architectural dimensions---depth, width, attention heads, and MLP intermediate size---rather than being limited to a single axis. As demonstrated in Table~\ref{tab:pruning_comparison}, methods like ShortGPT focus exclusively on depth pruning, while Wanda and SparseGPT target unstructured weight sparsity. In contrast, Minitron's multi-axis approach enables fine-grained control over the compression-performance trade-off, allowing us to identify an optimal ``Golden Ratio'' configuration (EXP\_010) that balances layer removal with intermediate dimension reduction. This hybrid flexibility proved essential for preserving the complex morphological and syntactic structures inherent in Polish while achieving our target parameter reduction of over 33\%.

Second, this work was conducted in direct \textbf{collaboration with NVIDIA}, which has open-sourced the Minitron tooling within the NeMo Framework and Model Optimizer \cite{nemo_minitron} in order to conduct sensitivity-based pruning and logit-matching distillation. This partnership enabled us to leverage production-grade infrastructure (NVIDIA DGX Cloud Lepton with H200 GPUs) and battle-tested compression pipelines that have been validated on models up to 15B parameters. Critically, our broader objective extends beyond a single model: we aim to establish a \textbf{reproducible blueprint for creating high-quality, efficient language models for European languages}, where computational resources are often more constrained than in English-centric research ecosystems. By demonstrating that Minitron-style compression can recover over 90\% of a teacher model's performance using less than 3\% of the original pre-training compute, we validate a pathway that reduces both the financial and environmental costs of LLM development. For less-represented languages like Polish, Czech, or Hungarian, this approach makes state-of-the-art NLP accessible without requiring the multi-million-dollar budgets typically associated with training frontier models from scratch.

\subsection{Positioning of Our Work}

The \textbf{approach} utilized in \cite{muralidharan2024compact} involves a sophisticated distillation pipeline where the Nemotron-4 15B  model \cite{parmar2024nemotron} serves as the teacher of a 4B student. The process begins with structured pruning---specifically targeting both width (hidden dimension, intermediate hidden dimension and attention heads) and depth (number of layers)---to derive the student architecture. This is followed by logit-based Knowledge Distillation (KD), which minimizes the Kullback--Leibler (KL) divergence between the student's and teacher's output distributions. Unlike standard cross-entropy training, this method allows the smaller model to inherit the teacher's nuanced probability distributions, effectively compressing its reasoning capabilities. Furthermore, the pipeline integrates a supervised fine-tuning (SFT) stage using high-quality synthetic data and Preference Alignment (DPO) \cite{rafailov2024direct} to refine the model's instruction-following accuracy.

The \textbf{results} of this methodology demonstrate a significant shift in the efficiency-to-performance ratio. Nemotron-Mini-4B-Instruct, despite its smaller footprint, outperformed several larger state-of-the-art models, including Llama-3-8B and Mistral-7B, across critical benchmarks. Notably, it achieved superior scores in mathematical reasoning (GSM8K) and coding tasks (HumanEval), proving that a distilled 4B model can rival or exceed the performance of 8B-class models when trained with high-fidelity teacher signals.

This work builds upon the Bielik-11B-v3.0 architecture \cite{ociepa2025bielik11bv3multilingual,ociepa2025bielik11bv2technical,Ociepa_Flis_Wrobel_Gwozdziej_Kinas_2025}, adapting these distillation and pruning insights for local deployment on consumer-grade hardware like the NVIDIA RTX 4090/5090 GPU series. By focusing on the logit-distillation patterns established in \cite{nvidia-developer-blogpost}, we aim to preserve the complex Polish-language reasoning of the larger Bielik variants within a more hardware-accessible parameter count.

\section{Methodology}

The transformation of \textbf{Bielik-11B-v3.0} into the compact yet high-performing \textbf{Bielik-Minitron-7B} model was carried out using a principled two-stage compression strategy followed by multi-level fine-tuning and alignment processes. Our methodology was inspired by recent advances in structured pruning and post-pruning knowledge recovery as demonstrated in large-scale transformer compression studies, particularly those employing NVIDIA Model Optimizer, NeMo, and Minitron-style distillation pipelines. The overarching objective was to achieve substantial parameter and latency reduction while preserving reasoning capability, distributional fidelity, and downstream task performance.

\subsection{Stage I: Structured Pruning via Importance-Aware Sensitivity Analysis}

We adopted \emph{structured pruning} as the primary compression mechanism, prioritizing hardware-efficient model reductions over sparsity-only approaches. Unlike unstructured pruning, which removes individual weights and often yields limited real-world speedups, structured pruning removes entire architectural components (e.g., transformer layers, attention heads, and feed-forward hidden dimension), ensuring compatibility with accelerator-optimized inference runtimes.

\paragraph{Pruning methods}
Model pruning is a fundamental optimization technique designed to address the computational inefficiencies of over-parameterized Large Language Models (LLMs). While increasing parameter counts typically enhances a model's functional capacity and knowledge density, it imposes significant costs in terms of inference latency and hardware requirements. Pruning aims to mitigate these drawbacks by strategically removing redundant parameters, thereby optimizing the model for deployment without substantial loss in output quality.

Structured pruning within this framework is generally categorized into three approaches:
\begin{itemize}
    \item \textbf{Depth pruning} involves the removal of entire transformer blocks or layers from the network. This approach significantly reduces the sequential complexity of the model, directly improving inference throughput.
    \item \textbf{Width pruning} focuses on reducing the internal dimensionality of the model by removing individual neurons, attention heads, or MLP intermediate dimensions. This results in a "narrower" architecture that reduces the memory footprint and per-layer computation.
    \item \textbf{Hybrid pruning} integrates both depth and width reduction, simultaneously scaling down the model across multiple architectural dimensions.
\end{itemize}

While independent application of depth or width pruning can yield efficient models, recent empirical evidence suggests that a hybrid approach — optimizing both dimensions in tandem — often produces student models that outperform those derived from single-dimensional pruning at equivalent parameter counts \cite{muralidharan2024compact}.

\paragraph{Pruning methodology}
To minimize performance degradation, the framework \cite{nvidia_nemo} identifies the least critical components using \textbf{activation-based importance estimation} on a calibration dataset. By performing forward passes on a representative subset of data, the framework collects activation statistics for each architectural unit---attention heads, MLP neurons, and embedding channels---and computes importance scores based on activation magnitudes. Components with consistently low activation magnitudes are identified as having minimal influence on the model's representational capacity and are therefore candidates for removal. This purely forward-pass approach avoids the computational overhead of gradient-based methods, enabling rapid architectural optimization without the need for extensive retraining.

\paragraph{Pruning candidates}
A comprehensive set of pruning configurations was established based on best practices from the Minitron approach \cite{sreenivas2024llm, muralidharan2024compact}, NVIDIA technical guidelines \cite{pruning_guidelines}, and preliminary empirical observations. These candidates were designed to explore the performance trade-offs between different architectural dimensions--specifically depth (layer count) and width (hidden and intermediate sizes)--while targeting a parameter reduction of approximately 25\% to 35\% from the base model. The detailed architectural specifications for each experimental scenario are summarized in Table~\ref{tab:pruning_scenarios}. Values which are not explicitly defined in the table remain as in the original model.

\begin{table*}[h]
    \centering
    \begin{tabular}{lcccc}
        \toprule
        \textbf{Experiment ID} & \textbf{Hidden size} & \textbf{Intermediate size} & \textbf{Layers} & \textbf{Total Parameters [B]} \\
        \midrule
        Original model & 4096 & 14336 & 50 & 11.04 \\
        \midrule
        EXP\_001  & 3072 & ---   & --- & 8.28 \\
        EXP\_002  & ---  & 9216  & --- & 7.90 \\
        EXP\_003  & ---  & ---   & 36  & 7.99 \\
        EXP\_004  & ---  & 12288 & 40  & 7.85 \\
        EXP\_005  & ---  & 10240 & --- & 8.52 \\
        EXP\_006  & ---  & 11264 & 44  & 8.07 \\
        EXP\_007  & 3648 & 12800 & 44  & 7.92 \\
        EXP\_008  & ---  & ---   & 32  & 7.11 \\
        EXP\_009  & ---  & 8192  & --- & 7.26 \\
        EXP\_010  & ---  & 11264 & 40  & 7.35 \\
        \bottomrule
    \end{tabular}
    \caption{Model pruning scenarios.}
    \label{tab:pruning_scenarios}
\end{table*}

\paragraph{Importance Estimation}
To identify pruning candidates, the framework utilizes a purely activation-based importance estimation strategy that requires only forward propagation passes. This approach assesses the significance of architectural units—specifically attention heads, MLP neurons, and embedding channels—by analyzing the magnitudes of their activations on a small calibration dataset. Specifically, for width-based pruning, importance scores are computed by calculating the $\ell_2$-norm across the batch dimension and the mean across the sequence length for the activations of each component. This criterion identifies units with minimal activation magnitudes as having the least influence on the model's representational capacity. By ranking components according to these scores, the framework can perform structured pruning efficiently, avoiding the computational overhead associated with gradient-based or loss-sensitivity calculations.

\paragraph{Granularity of Pruning}
Importance scores were computed across multiple structural levels to identify redundancies at different scales of the transformer architecture: 

\begin{itemize} 
\item \textbf{Transformer Depth:} Entire residual blocks were evaluated for their contribution to the residual stream, allowing for the removal of redundant layers. 
\item \textbf{Multi-Head Attention (MHA):} Individual attention heads were ranked, enabling the reduction of the model's width while preserving the most critical heads for capturing long-range dependencies. 
\item \textbf{Feed-Forward Networks (FFN):} The intermediate hidden dimensions of the MLP layers were pruned, targeting the components with the lowest activation magnitudes. 
\end{itemize}

This multi-granular analysis allowed us to jointly optimize depth and width reductions while maintaining architectural balance.

\paragraph{Final Pruned Configuration}
Based on a comprehensive analysis of sensitivity rankings and empirical validation across the candidate pool, the configuration corresponding to \textbf{EXP\_010} was selected as the optimal architecture. This strategy reduced the model depth from 50 to 40 transformer layers and downscaled the FFN intermediate dimension from 14,336 to 11,264. The resulting student model, \textbf{Bielik-Minitron-7B}, comprises approximately 7.35B parameters—representing a 33.4\% reduction in total parameter count from the 11.04B baseline. Crucially, this configuration preserved the original hidden dimension ($d_{model} = 4096$) and the multi-head attention topology, including the Rotary Positional Embedding (RoPE) scheme. By maintaining these structural invariants, we were able to initialize the student directly with the most significant weights from the teacher, providing a high-fidelity foundation for the subsequent knowledge recovery phase. As our experiments did not include pruning the attention heads, this dimension remains a possibility for future work.

\subsection{Stage II: Knowledge Distillation for Post-Pruning Recovery}

While structured pruning delivers efficient hardware-level compression, the abrupt removal of weights and layers inevitably introduces representational damage and disrupts the internal feature maps of the model. To mitigate these effects, we employed \emph{knowledge distillation} (KD) as a post-pruning recovery mechanism. Unlike standard pre-training, which relies solely on ground-truth token labels, this stage enables the student model to re-approximate the complex functional behavior of the original Bielik-11B-v3.0 network by learning from its full output distribution.

By minimizing the divergence between the teacher's and the student's logits, the pruned model inherits the "dark knowledge" of the larger network—specifically the nuanced probabilities assigned to incorrect or alternative tokens. This process allows the 7.35B-parameter student to recover linguistic reasoning and distributional fidelity that would otherwise require significantly more data and compute if trained from scratch.

\paragraph{Teacher--Student Setup}
We utilized the \textbf{NVIDIA NeMo Framework (v24.09)} to facilitate large-scale, distributed distillation. In this configuration, the original \textbf{Bielik-11B-v3.0} model served as a frozen \emph{teacher}, providing a high-dimensional reference signal. The pruned \textbf{Bielik-Minitron-7B} model acted as the \emph{student}, having been initialized directly with the surviving weights of the teacher to preserve as much pre-existing knowledge as possible. The student was trained to minimize the discrepancy between its output logits and those of the teacher over a diverse corpus of unlabeled Polish and multi-lingual text. This logit-based alignment ensures that the student inherits not only the teacher's top-1 predictions but also its nuanced reasoning patterns and underlying calibration properties.

\paragraph{Distillation Objective}
Following the best practices established in \cite{muralidharan2024compact}, the training objective employed a logit-only forward KL divergence loss:
\begin{equation}
    \mathcal{L} = \text{KL}\left(\sigma(\mathbf{z}_t/T) \| \sigma(\mathbf{z}_s/T)\right)
\end{equation}
where $\mathbf{z}_t$ and $\mathbf{z}_s$ denote the teacher and student logits, $\sigma$ is the softmax function, and $T$ is the temperature hyperparameter. Crucially, the standard cross-entropy loss against ground-truth labels was not used during distillation---the student learned exclusively from the teacher's output distribution. This design choice follows the finding that logit-only distillation provides superior recovery when depth reduction is moderate, as in our configuration (50 $\rightarrow$ 40 layers). By focusing solely on the KL divergence between the full probability distributions, the student inherits not only the teacher's top-1 predictions but also the nuanced inter-token relationships and confidence calibration encoded in the long tail of the distribution.

\paragraph{Temperature Scaling}
Temperature scaling was applied to the softmax logits during the distillation process. The temperature $T$ flattens the teacher's probability distributions, amplifying signals from low-probability tokens that reside in the long tails of the distribution. By increasing the entropy of the target distribution, this scaling enabled the student to capture nuanced decision boundaries and subtle linguistic dependencies that are often suppressed during standard hard-label training. This proved instrumental in preserving the teacher's calibration, particularly for the complex morphological structures inherent in the Polish language.

\subsection{Alignment Objectives and Outcomes}
The integrated pruning-and-distillation pipeline was engineered to address the inherent trade-offs between model compression and linguistic capability, focusing on three primary objectives:
\begin{enumerate}
    \item \textbf{Efficiency:} Achieving a significant reduction in parameter count and inference latency without introducing architectural fragmentation, thereby ensuring compatibility with standard optimization toolkits like TensorRT-LLM.
    \item \textbf{Fidelity:} Preserving the teacher model's representational depth, reasoning logic, and nuanced output distribution, particularly for the specific syntactic requirements of the Polish language.
    \item \textbf{Stability:} Maintaining training and inference robustness after aggressive structural reduction, preventing the "catastrophic forgetting" or divergence often associated with high-compression regimes.
\end{enumerate}

Empirically, the distilled \textbf{Bielik-Minitron-7B} successfully recovered the vast majority of the performance margin lost during the initial pruning phase. Our internal evaluation benchmarks indicate that the student model achieves near-parity with the \textbf{Bielik-11B-v3.0} baseline, while delivering a substantial increase in throughput and a reduced memory footprint. These results validate that importance-aware structured pruning, when coupled with high-temperature logit distillation, provides a highly effective and computationally efficient pathway for producing compact, deployment-ready language models for localized markets.

\subsection{Post-Distillation Alignment Pipeline}
To transition the distilled base model into a production-ready assistant, the 7.35B student underwent a rigorous, multi-stage alignment protocol. This pipeline was designed to mirror the treatment of the \textbf{Bielik-11B-v3.0} \cite{ociepa2025bielik11bv3multilingual} flagship, ensuring that the efficiency gains of pruning did not come at the expense of instruction-following precision or safety.

\begin{enumerate}
    \item \textbf{Supervised Fine-Tuning (SFT):} The model was fine-tuned on a curated set of high-quality Polish and English instruction-following pairs. This stage established the primary conversational interface and aligned the model with specific linguistic nuances and formatting requirements of the Polish market. The model was trained for 3 epochs on approximately 20 million instructions and dialogue samples, using a maximum sequence length of 32{,}768 tokens.
    
    \item \textbf{Preference Alignment (DPO-P)\cite{pal2024smaugfixingfailuremodes}:} We employed \emph{Direct Preference Optimization} using a "Positive" (DPO-P) variant. By focusing on stabilizing the policy gradient and emphasizing high-reward trajectories, this stage improved the model's adherence to human preferences and significantly reduced the likelihood of generating harmful or irrelevant content. The training was conducted for 3 epochs on a dataset of 114{,}000 preference-labeled samples, with a maximum sequence length of 32{,}768 tokens.
    
    \item \textbf{Reinforcement Learning (GRPO)\cite{shao2024deepseek}:} To bridge the reasoning gap often found in smaller models, we integrated \emph{Group Relative Policy Optimization}. By utilizing verifiable reward functions for STEM, mathematical, and logical tasks, GRPO allowed the model to explore "thought chains" and self-correct without the computational overhead of a separate critic network. This stage was trained on 143{,}000 task-specific samples.
\end{enumerate}

\section{Systematic Architecture Search}
To identify the "Golden Ratio" of parameter reduction—defined as the optimal balance between computational throughput and representational capacity—we conducted an extensive search across 10 distinct pruning configurations. Our empirical findings revealed a non-linear relationship between structural reduction and model stability. Aggressive width-only pruning (e.g., \textbf{EXP\_009}, 34.2\% reduction) led to significant training instability and gradient spikes during the initial phases of distillation. Conversely, conservative pruning (e.g., \textbf{EXP\_006}, 26.9\% reduction) failed to meet our predefined efficiency targets for local deployment. \textbf{EXP\_010} was ultimately selected as the optimal production candidate, providing the most robust recovery curve and the best hardware-utilization profile.

\begin{table*}[h]
    \centering
    \begin{tabular}{lcccc}
        \toprule
        \textbf{Experiment} & \textbf{Parameters} & \textbf{Reduction} & \textbf{Structural Modifications} & \textbf{Status} \\
        \midrule
        Original (11B-v3) & 11.04B & --- & Baseline Configuration & Baseline \\
        EXP\_006 & 8.07B & 26.9\% & Layers: 50 $\rightarrow$ 44, Intermediate: 14336 $\rightarrow$ 11264 & Sub-optimal \\
        EXP\_009 & 7.26B & 34.2\% & Intermediate: 14336 $\rightarrow$ 8192 & Unstable \\
        \textbf{EXP\_010} & \textbf{7.35B} & \textbf{33.4\%} & \textbf{Layers: 40, Intermediate: 11264} & \textbf{Selected} \\
        \bottomrule
    \end{tabular}
    \caption{Systematic Search: Comparison of Key Pruning Candidates.}
    \label{tab:arch-search}
\end{table*}

\section{Experimental Setup}

\subsection{Data and Hardware}
The training process utilized the \textbf{Bielik Dataset}, a curated Polish-English corpus comprising 8.0M high-quality samples (raw text samples for pretraining). This dataset was specifically balanced to maintain the model's multilingual proficiency while deepening its grasp of Polish idiomatic and technical structures.

Computational workloads focused on pruning and distillation processes were executed on a cluster comprising two nodes totaling \textbf{16 NVIDIA H200 GPUs} (via DGX Cloud Lepton). The H200's \textbf{141GB of HBM3e} memory proved to be a critical enabler for the distillation pipeline. With a total memory bandwidth of 4.8 TB/s and expanded capacity, the hardware allowed both the 11.04B teacher and the 7.35B student models to reside in the GPU VRAM simultaneously. This high-residency configuration eliminated the need for activation offloading or intensive tensor parallelism, significantly reducing inter-GPU communication overhead and facilitating a highly efficient, high-throughput distillation process.

The remaining post-training processes were conducted as usual for the Bielik project at ACK Cyfronet AGH. This part included the training pipeline which was used for the creation of the Bielik-11B-v3.0 model, further described in \cite{ociepa2025bielik11bv3multilingual}. 

\subsection{Training Dynamics}

The distillation of each architectural candidate was orchestrated over a period of 48--72 hours, with the comprehensive experimental search and validation cycle encompassing approximately three weeks. To govern the optimization trajectory, we employed the \textbf{AdamW optimizer} coupled with a \textbf{cosine annealing scheduler}; the learning rate was initialized at a peak of $1.5 \times 10^{-4}$ and was systematically attenuated to a terminal value of $1.5 \times 10^{-5}$.

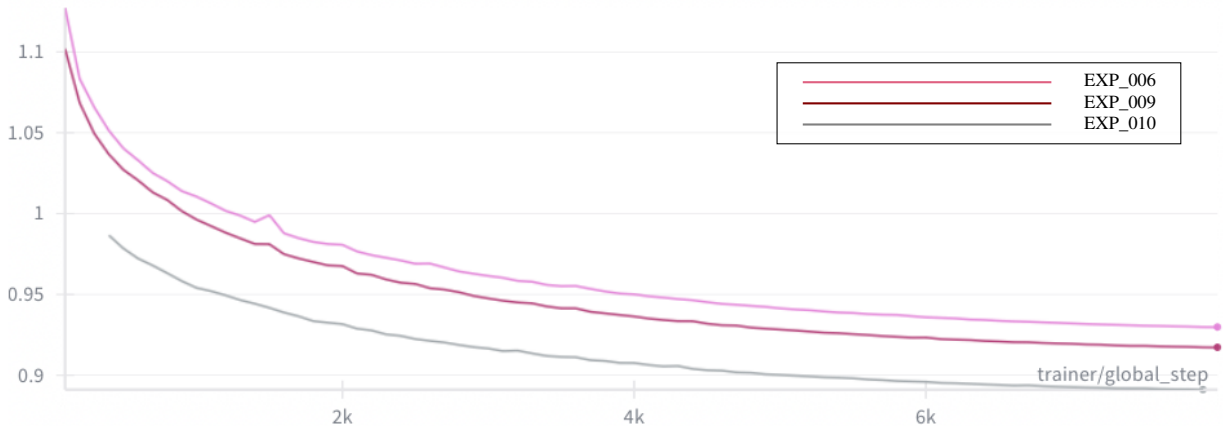
\begin{figure}[ht]
    \centering
    \begin{tikzpicture}
        \node[anchor=south west,inner sep=0] (image) at (0,0) {\includegraphics[width=1\linewidth]{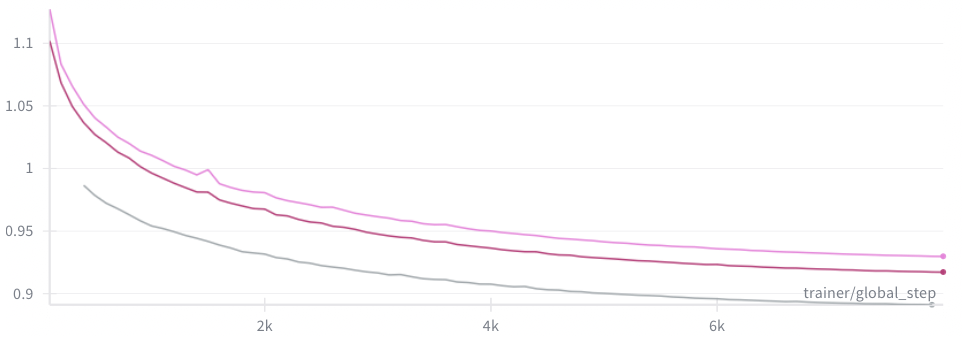}};
        
        \begin{scope}[x={(image.south east)},y={(image.north west)}]
            \node [draw, fill=white, fill opacity=0.8, text opacity=1, anchor=north east, font=\scriptsize] at (0.95,0.85) {
                \begin{tabular}{ll}
                \protect\tikz[baseline=-0.5ex]\protect\draw[pink!50!purple,thick] (0,0) -- (0.2,0); & EXP\_006 \\
                \protect\tikz[baseline=-0.5ex]\protect\draw[red!50!black,thick] (0,0) -- (0.2, 0); & EXP\_009 \\
                \protect\tikz[baseline=-0.5ex]\protect\draw[gray,thick] (0,0) -- (0.2,0); & EXP\_010 \\
                \end{tabular}
            };
        \end{scope}
    \end{tikzpicture}
    \caption{Training convergence profiles of distillation loss. The legend indicates the three architectural variants evaluated during the search phase - EXP\_006, EXP\_009 and EXP\_010.}
    \label{fig:loss_convergence}
\end{figure}

As evidenced by our telemetry (Figure~\ref{fig:loss_convergence}), the global distillation training loss exhibited a seamless and stable convergence across all variants. The loss objective exhibits a smooth, monotonic decay from an initial $\sim$1.12 to a converged steady state near $\sim$0.89 over 8,000 optimization steps, signifying high numerical stability during the distillation phase. By leveraging the \textbf{NVIDIA NeMo} implementation of logit-matching, we maintained a sustained \textbf{90\% GPU utilization} across the two-node H200 cluster. This high level of hardware saturation, combined with the lack of memory-offloading bottlenecks, resulted in a significant reduction in total wall-clock time compared to traditional distillation setups on previous-generation hardware.

\section{Results \& Evaluation}
The evaluation of \textbf{Bielik-Minitron-7B} focused on its ability to replicate the teacher's performance across the most challenging Polish linguistic and logical benchmarks. Our primary candidate, \textbf{EXP\_010}, successfully retained \textbf{90.1\%} of the \textbf{Bielik-11B-v3.0} baseline's performance. As shown in Table~\ref{tab:results-recovery-detailed}, the model exhibited exceptional recovery in semantic and affective tasks, while maintaining a high degree of generative quality.

\begin{table*}[h]
    \centering
    \begin{tabular}{lcc}
        \toprule
        \textbf{Task Category} & \textbf{Recovery \%} & \textbf{Evaluation Metric} \\
        \midrule
        \textbf{Overall Aggregate Score} & \textbf{90.1\%} & \textbf{Avg. Benchmark Accuracy} \\
        Open PL LLM Leaderboard & 94.7\%  & Average Score \\
        Polish EQ-Bench & 90.0\%  & Average Score \\
        CPTUB & 90.6\%  & Average Score \\
        Polish Medical Leaderboard & 88.3\%  & Accuracy \\
        INCLUDE-base-44 & 88.6\%  & Regional Knowledge \\
        Belebele & 94.0\%  & Reading Comprehension \\
        FLORES Machine Translation & 80.8\%  & BLEU Score \\
        EuroEval & 91.4\%  & Average Score \\
        BFCL & 92.3\% & Average Score \\
        \bottomrule
    \end{tabular}
    \caption{Performance Recovery Analysis: Student (7.35B) vs. Teacher (11.04B).}
    \label{tab:results-recovery-detailed}
\end{table*}

\subsection{Open PL LLM Leaderboard}
Open PL LLM Leaderboard is a benchmark inspired by Open LLM Leaderboard \cite{open-llm-leaderboard} and described in detail in the Bielik v0.1 paper \cite{Ociepa_Flis_Wrobel_Gwozdziej_Kinas_2025}. Evaluation tasks consist of various NLP tasks testing sentiment analysis, categorization or text classification explicitly in Polish language. Some of the tasks included are the likes of \textbf{polemo2, klej-ner, dyk, polqa} or \textbf{poquad}. Scores presented are average across all tasks in 5-shot evaluation manner. 

The performance of \textbf{Bielik-Minitron-7B-v3.0-Instruct} on the Open PL LLM Leaderboard (Table~\ref{tab:open-pl-llm-instruct}) validates the efficiency of the pruning and knowledge recovery pipeline. Despite a 33.4\% reduction in parameter count, the 7.35B student model retains approximately 94.7\% of the performance of its parent, \textbf{Bielik-11B-v3.0-Instruct} (65.93). Notably, it represents a massive generational leap over the original \textbf{Bielik-7B-Instruct-v0.1} (44.70), demonstrating that distilling a high-capacity 11B teacher is significantly more effective than traditional training at the 7B scale.

When compared to external competitors, \textbf{Bielik-Minitron-7B} exhibits best-in-class efficiency. It significantly outperforms direct rivals of similar scale, such as \textbf{Qwen2.5-7B-Instruct} (54.93) and \textbf{Mistral-7B-v0.3} (47.74), leading by 7.53 and 14.72 points respectively. The model achieves parity with or exceeds the performance of much larger architectures, including \textbf{Qwen3-14B} (62.24), \textbf{gemma-3-12b-it} (62.20), and even \textbf{Qwen2.5-32B-Instruct} (61.21). 

It ranks as one of the most capable models under 10B parameters on the leaderboard, trailing the 14.7B \textbf{phi-4} (62.57) by only 0.11 points. These results suggest that the Minitron methodology effectively preserves the linguistic nuances and capabilities of the teacher model while drastically reducing the computational footprint.

\begin{table*}[t]
    \centering
    \small
    \begin{tabular}{lrr}
    \toprule
    \textbf{Model} & \textbf{Parameters (B)} & \textbf{Average} \\
    \midrule
    Mistral-Large-Instruct-2411 & 123.0 & 69.84 \\
    Meta-Llama-3.1-405B-Instruct-FP8 & 405.0 & 69.44 \\
    Mistral-Large-Instruct-2407 & 123.0 & 69.11 \\
    Qwen2.5-72B-Instruct & 72.7 & 67.92 \\
    QwQ-32B-Preview & 32.8 & 67.01 \\
    Llama-3.3-70B-Instruct & 70.6 & 66.40 \\
    \textbf{Bielik-11B-v3.0-Instruct} & \textbf{11.2} & \textbf{65.93} \\
    Qwen2-72B-Instruct & 72.7 & 65.87 \\
    \underline{Bielik-11B-v2.3-Instruct} & \underline{11.2} & \underline{65.71} \\
    \underline{Bielik-11B-v2.2-Instruct} & \underline{11.2} & \underline{65.57} \\
    Meta-Llama-3.1-70B-Instruct & 70.6 & 65.49 \\
    \underline{Bielik-11B-v2.1-Instruct} & \underline{11.2} & \underline{65.45} \\
    Mixtral-8x22B-Instruct-v0.1 & 141.0 & 65.23 \\
    \underline{Bielik-11B-v2.0-Instruct} & \underline{11.2} & \underline{64.98} \\
    Meta-Llama-3-70B-Instruct & 70.6 & 64.45 \\
    \underline{Bielik-11B-v2.6-Instruct} & \underline{11.2} & \underline{64.26} \\
    Qwen3-32B & 32.8 & 64.24 \\
    Llama-4-Scout-17B-16E-Instruct & 109.0 & 64.21 \\
    \underline{Bielik-11B-v2.5-Instruct} & \underline{11.2} & \underline{63.95} \\
    Mistral-Small-24B-Instruct-2501 & 24.0 & 62.97 \\
    phi-4 & 14.7 & 62.57 \\
    \textbf{Bielik-Minitron-7B-v3.0-Instruct} & \textbf{7.35} & \textbf{62.46} \\
    Qwen3-14B & 14.8 & 62.24 \\
    gemma-3-12b-it & 12.0 & 62.20 \\
    Mistral-Small-Instruct-2409 & 22.2 & 61.41 \\
    Qwen2.5-32B-Instruct & 32.8 & 61.21 \\
    Qwen2.5-14B-Instruct & 14.8 & 59.91 \\
    aya-23-35B & 35.0 & 56.37 \\
    \underline{Bielik-4.5B-v3.0-Instruct} & \underline{4.8} & \underline{56.13} \\
    gemma-3-27b-it & 27.0 & 55.92 \\
    Qwen3-8B & 8.2 & 55.78 \\
    Qwen3-4B & 4.0 & 55.49 \\
    Mistral-Nemo-Instruct-2407 & 12.2 & 55.27 \\
    EuroLLM-22B-Instruct-Preview & 22.0 & 55.17 \\
    Qwen2.5-7B-Instruct & 7.6 & 54.93 \\
    EuroLLM-9B-Instruct & 9.0 & 50.07 \\
    GaMS-9B-Instruct & 9.0 & 48.78 \\
    Mistral-7B-Instruct-v0.3 & 7.2 & 47.74 \\
    Apertus-8B-Instruct-2509 & 8.0 & 47.27 \\
    Mistral-7B-Instruct-v0.2 & 7.2 & 45.95 \\
    \underline{Bielik-7B-Instruct-v0.1} & \underline{7.2} & \underline{44.70} \\
    gemma-2-9b-it & 9.0 & 42.12 \\  
    Qwen2.5-3B-Instruct & 3.0 & 41.23 \\
    Mistral-7B-Instruct-v0.1 &	7.0 &	33.11 \\
    Qwen2.5-1.5B-Instruct & 1.5 & 31.89 \\
    \bottomrule
    \end{tabular}
    \caption{Open PL LLM Leaderboard results for instruction-tuned models (5-shot evaluation)}
    \label{tab:open-pl-llm-instruct}
    \end{table*}

\subsection{Polish EQ-Bench}
The evaluation in the \textbf{Polish EQ-Bench} (Table~\ref{tab:pl-eq-bench}) assesses the models' capacity for emotional understanding and social reasoning in a Polish context. \textbf{Bielik-Minitron-7B-v3.0-Instruct} achieved a score of \textbf{64.09}, placing it firmly in the mid-tier of the leaderboard despite being significantly smaller than its competitors.
    
\begin{table*}[t]
    \centering
    \small
    \begin{tabular}{lrc}
    \toprule
    \textbf{Model} & \textbf{Parameters (B)} & \textbf{Score} \\
    \midrule
    Mistral-Large-Instruct-2407$^{\dagger}$ & 123.0 & 78.07 \\
    Mistral-Large-Instruct-2411$^{\dagger}$ & 123.0 & 77.29 \\
    Meta-Llama-3.1-405B-Instruct-FP8 & 405.0 & 77.23 \\
    gpt-4o-2024-08-06 & Unknown & 75.15 \\
    gpt-4-turbo-2024-04-09 & Unknown & 74.59 \\
    \underline{Bielik-11B-v2.6-Instruct} & \underline{11.2} & \underline{73.8} \\
    DeepSeek-V3-0324 & 685.0 & 73.46 \\
    Mistral-Small-Instruct-2409 & 22.2 & 72.85 \\
    Llama-PLLuM-70B-chat & 70.6 & 72.56 \\
    Meta-Llama-3.1-70B-Instruct & 70.6 & 72.53 \\
    \underline{Bielik-11B-v2.5-Instruct} & \underline{11.2} & \underline{72.00} \\
    Qwen2-72B-Instruct & 72.7 & 71.23 \\
    Meta-Llama-3-70B-Instruct & 70.6 & 71.21 \\
    \textbf{Bielik-11B-v3.0-Instruct} & \textbf{11.2} & \textbf{71.20} \\

    gpt-4o-mini-2024-07-18 & Unknown & 71.15 \\
    Qwen2.5-32B-Instruct & 32.8 & 71.15 \\
    \underline{Bielik-11B-v2.3-Instruct} & \underline{11.2} & \underline{70.86} \\
    Llama-3.3-70B-Instruct & 70.6 & 70.73 \\
    Llama-PLLuM-70B-instruct & 70.6 & 69.99 \\
    WizardLM-2-8x22B & 141.0 & 69.56 \\
    Qwen2.5-14B-Instruct & 14.8 & 69.17 \\
    \underline{Bielik-11B-v2.2-Instruct} & \underline{11.2} & \underline{69.05} \\
    \underline{Bielik-11B-v2.0-Instruct} & \underline{11.2} & \underline{68.24} \\
    \textbf{Bielik-Minitron-7B-v3.0-Instruct} & \textbf{7.35} & \textbf{64.09} \\
    glm-4-9b-chat & 9.0 & 61.79 \\
    Mistral-Nemo-Instruct-2407 & 12.2 & 61.76 \\
    \underline{Bielik-11B-v2.1-Instruct} & \underline{11.2} & \underline{60.07} \\
    pllum-12b-nc-chat-250715 & 12.2 & 55.20 \\
    EuroLLM-9B-Instruct & 9.2 & 54.10 \\
    \underline{Bielik-4.5B-v3.0-Instruct} & \underline{4.8} & \underline{53.58} \\
    PLLuM-12B-chat & 12.2 & 52.26 \\
    PLLuM-8x7B-nc-chat$^{\dagger}$ & 46.7 & 47.29 \\
    Llama-PLLuM-8B-chat & 8.0 & 46.20 \\
    PLLuM-8x7B-chat & 46.7 & 45.22 \\
    PLLuM-12B-nc-chat$^{\dagger}$ & 12.2 & 35.41 \\
    \bottomrule
    \multicolumn{3}{l}{$^{\dagger}$Models with a non-commercial license.} \\
    \end{tabular}
    \caption{Polish EQ-Bench results for various models.}
    \label{tab:pl-eq-bench}
    \end{table*}

The 7.35B student model maintains approximately 90\% of the emotional reasoning capabilities of the \textbf{Bielik-11B-v3.0-Instruct} parent (71.20). In particular, it outperforms the earlier 11B variant \textbf{Bielik-11B-v2.1} (60.07), suggesting that the knowledge recovery phase effectively transfers the refined social nuances of the v3.0 teacher. \textbf{Bielik-Minitron-7B} exceeds the scores of several larger models, including \textbf{Mistral-Nemo-Instruct-2407} (61.76, 12.2B) and the \textbf{PLLuM-12B-chat} (52.26).

While the model trails the \textbf{Mistral-Small-24B} and much larger 70B+ architectures, its ability to outperform \textbf{glm-4-9b-chat} (61.79) demonstrates that the pruning process preserves complex "soft skills" like emotional intensity prediction better than some larger, non-distilled models.

\subsection{Complex Polish Text Understanding Benchmark (CPTUB)}
The \textbf{CPTUB} results (Table~\ref{tab:cptub}) provide a granular look at the models' mastery of the Polish language in specialized categories, which is well known for its complex morphological and syntactic structures.

In the \textbf{Language Understanding} category, the 7.35B student model achieves a score of \textbf{3.83}, which is close to the 11B teacher's \textbf{3.91}. This indicates that the pruning process successfully preserved the core syntactic and grammatical capabilities of the larger model. \textbf{Bielik-Minitron-7B} outperforms several larger or international models, including \textbf{phi-4} (3.30) and \textbf{Qwen2.5-7B-Instruct} (3.07). It also matches the performance of \textbf{Llama-PLLuM-70B-instruct} (3.33), despite a nearly 10x difference in parameter count. The model shows particular resilience in \textbf{Sentiment} (3.72) and \textbf{Implicatures} (3.59). Although there is an expected drop in \textbf{Phraseology} and \textbf{Tricky Questions} compared to the 11B teacher, the student remains significantly more capable than the first generation of Bielik family -- \textbf{Bielik-7B-v0.1} (2.88).

\begin{table*}[t]
    \centering
    \small
    \begin{tabular}{lrcccccc}
    \toprule
    \textbf{Model} & \textbf{Params (B)} & \textbf{Overall} & \textbf{Implicatures} & \textbf{Senti-} & \textbf{Language} & \textbf{Phrase-} & \textbf{Tricky} \\
    & & \textbf{Average} & \textbf{Average} & \textbf{ment} & \textbf{Understanding} & \textbf{ology} & \textbf{Questions} \\
    \midrule
    gemini-2.0-flash-001 & Unknown & 4.29 & 4.39 & 4.52 & 4.32 & 4.34 & 3.99 \\
    DeepSeek-R1 & 685.0 & 4.14 & 4.14 & 4.49 & 4.35 & 3.60 & 4.12 \\
    gemini-2.0-flash-lite-001 & Unknown & 4.09 & 4.17 & 4.23 & 4.05 & 4.24 & 3.85 \\
    DeepSeek-V3-0324 & 685.0 & 4.03 & 4.03 & 4.36 & 4.20 & 3.54 & 4.02 \\
    Mistral-Large-Instruct-2411$^{\dagger}$ & 123.0 & 4.00 & 4.10 & 4.33 & 3.98 & 3.99 & 3.72 \\
    Qwen2.5-72B-Instruct & 72.7 & 3.95 & 3.99 & 4.08 & 3.97 & 3.93 & 3.81 \\
    Mistral-Large-Instruct-2407$^{\dagger}$ & 123.0 & 3.93 & 4.03 & 4.23 & 4.00 & 3.86 & 3.65 \\
    Llama-4-Maverick-17B-128E-Instruct & 402.0 & 3.93 & 3.99 & 4.39 & 4.11 & 3.48 & 3.76 \\
    gemma-3-27b-it & 27.4 & 3.81 & 3.90 & 3.88 & 3.79 & 4.03 & 3.53 \\
    Meta-Llama-3-70B-Instruct & 70.6 & 3.78 & 3.81 & 4.13 & 3.82 & 3.47 & 3.71 \\
    Qwen2.5-32B-Instruct & 32.8 & 3.75 & 3.80 & 3.81 & 3.57 & 4.04 & 3.59 \\
    Llama-4-Scout-17B-16E-Instruct & 109.0 & 3.75 & 3.94 & 4.10 & 3.81 & 3.90 & 3.19 \\
    \textbf{Bielik-11B-v3.0-Instruct} & \textbf{11.2} & \textbf{3.73} & \textbf{3.92} & \textbf{3.88} & \textbf{3.91} & \textbf{3.96} & \textbf{3.19} \\
    Mistral-Small-24B-Instruct-2501 & 23.6 & 3.71 & 3.80 & 3.91 & 3.60 & 3.88 & 3.45 \\
    pllum-12b-nc-chat-250715$^{\dagger}$ & 12.2 & 3.67 & 3.92 & 4.36 & 3.96 & 3.46 & 2.90 \\
    \underline{Bielik-11B-v2.6-Instruct} & \underline{11.2} & \underline{3.64} & \underline{3.82} & \underline{4.10} & \underline{3.94} & \underline{3.41} & \underline{3.10} \\
    Mixtral-8x22B-Instruct-v0.1 & 141.0 & 3.56 & 3.67 & 3.78 & 3.68 & 3.55 & 3.24 \\
    Qwen2.5-14B-Instruct & 14.8 & 3.55 & 3.62 & 3.91 & 3.57 & 3.37 & 3.34 \\
    Llama-PLLuM-70B-chat & 70.6 & 3.53 & 3.63 & 3.94 & 3.61 & 3.35 & 3.21 \\
    \textbf{Bielik-Minitron-7B-v3.0-Instruct} & \textbf{7.35} & \textbf{3.38} & \textbf{3.59} & \textbf{3.72} & \textbf{3.83} & \textbf{3.23} & \textbf{2.74} \\
    \underline{Bielik-4.5B-v3.0-Instruct} & \underline{4.8} & \underline{3.38} & \underline{3.68} & \underline{3.76} & \underline{3.61} & \underline{3.67} & \underline{2.46} \\
    Llama-PLLuM-70B-instruct & 70.6 & 3.33 & 3.56 & 3.78 & 3.63 & 3.26 & 2.63 \\
    phi-4 & 14.7 & 3.30 & 3.50 & 3.72 & 3.54 & 3.24 & 2.72 \\
    PLLuM-12B-chat & 12.2 & 3.14 & 3.32 & 3.32 & 3.21 & 3.43 & 2.59 \\
    PLLuM-8x7B-nc-instruct$^{\dagger}$ & 46.7 & 3.11 & 3.56 & 3.88 & 3.59 & 3.22 & 1.76 \\
    Qwen2.5-7B-Instruct & 7.62 & 3.07 & 3.23 & 3.56 & 3.03 & 3.10 & 2.58 \\
    EuroLLM-9B-Instruct & 9.0 & 3.15 & 3.28 & 3.37 & 3.30 & 3.17 & 2.75 \\
    PLLuM-8x7B-nc-chat$^{\dagger}$ & 46.7 & 3.03 & 3.44 & 3.76 & 3.48 & 3.08 & 1.80 \\
    Meta-Llama-3.1-8B-Instruct & 8.0 & 3.01 & 3.31 & 3.97 & 3.38 & 2.58 & 2.11 \\
    PLLuM-8x7B-chat & 46.7 & 3.01 & 3.41 & 3.44 & 3.45 & 3.35 & 1.78 \\
    Meta-Llama-3-8B-Instruct & 8.0 & 3.00 & 3.17 & 3.33 & 3.15 & 3.04 & 2.48 \\
    Llama-PLLuM-8B-chat & 8.0 & 2.92 & 3.14 & 3.13 & 2.93 & 3.36 & 2.25 \\
    \underline{Bielik-7B-Instruct-v0.1} & \underline{7.2} & \underline{2.88} & \underline{3.13} & \underline{3.59} & \underline{3.48} & \underline{2.32} & \underline{2.16} \\
    \bottomrule
    \multicolumn{8}{l}{$^{\dagger}$Models with a non-commercial license.} \\
    \end{tabular}
    \caption{Complex Polish Text Understanding Benchmark (CPTUB) results across different evaluation categories}
    \label{tab:cptub}
\end{table*}

\subsection{Polish Medical Leaderboard}
The evaluation on the \textbf{Polish Medical Leaderboard} (Table~\ref{tab:medical-leaderboard}) tests the models' domain-specific knowledge using questions from the \textbf{Medical Final Examination (LEK)} and the \textbf{State Specialization Exam (PES)}. \textbf{Bielik-Minitron-7B-v3.0-Instruct} achieved an average score of \textbf{44.36\%}, demonstrating a high degree of retention of specialized knowledge.

The 7.35B student model performs on par with several 11.2B predecessors, including \textbf{Bielik-11B-v2.6-Instruct} (44.88\%) and \textbf{Bielik-11B-v2.5-Instruct} (44.85\%). Crucially, it outperforms the v2.3, v2.2, and v2.1 generations of the 11B model, suggesting that the v3.0 teacher passed down a more refined medical knowledge base that survived the pruning process.

\textbf{Bielik-Minitron-7B} exceeds the performance of notably larger models such as \textbf{Mistral-Small-Instruct-2409} (43.60\%, 22.2B) and \textbf{Mistral-Nemo-Instruct-2407} (40.36\%, 12.2B). It also maintains a lead over established small models like \textbf{Meta-Llama-3.1-8B-Instruct} (40.60\%). Within its immediate size category, the model shows superior medical reasoning compared to \textbf{Qwen2.5-7B-Instruct} (42.69\%) and provides a massive improvement (+14.62 points) over the previous \textbf{Bielik-7B-v0.1} (29.74\%).

Although there is a 5.85-point gap between the student and the \textbf{Bielik-11B-v3.0-Instruct} teacher (50.21\%), the student remains highly competitive with nearly all models in the 10B--25B parameter range.

\begin{table*}[t]
    \centering
    \small
    \begin{tabular}{lrc}
    \toprule
    \textbf{Model} & \textbf{Parameters (B)} & \textbf{Average (\%)} \\
    \midrule
    Meta-Llama-3.1-405B-Instruct-FP8 & 405.0 & 69.20 \\
    Mistral-Large-Instruct-2407$^{\dagger}$ & 123.0 & 64.28 \\
    Qwen2.5-72B-Instruct & 72.7 & 63.89 \\
    Meta-Llama-3.1-70B-Instruct & 70.6 & 61.75 \\
    Qwen2-72B-Instruct & 72.7 & 61.35 \\
    Meta-Llama-3-70B-Instruct & 70.6 & 57.51 \\
    Qwen2.5-32B & 32.8 & 55.69 \\
    Qwen2.5-32B-Instruct & 32.8 & 54.52 \\
    \textbf{Bielik-11B-v3.0-Instruct} & \textbf{11.2} & \textbf{50.21} \\
    Qwen2.5-14B-Instruct & 14.8 & 49.60 \\
    \textbf{Bielik-11B-v3-Base-20250730} & \textbf{11.2} & \textbf{45.86} \\
    \underline{Bielik-11B-v2.6-Instruct} & \underline{11.2} & \underline{44.88} \\
    \underline{Bielik-11B-v2.5-Instruct} & \underline{11.2} & \underline{44.85} \\
    GLM-4-9b-chat & 9.0 & 44.54 \\
    \textbf{Bielik-Minitron-7B-v3.0-Instruct} & \textbf{7.35} & \textbf{44.36} \\
    Mistral-Small-Instruct-2409 & 22.2 & 43.60 \\
    \underline{Bielik-4.5B-v3.0-Instruct} & \underline{4.8} & \underline{43.55} \\
    \underline{Bielik-11B-v2.3-Instruct} & \underline{11.2} & \underline{43.26} \\
    \underline{Bielik-11B-v2.1-Instruct} & \underline{11.2} & \underline{43.16} \\
    \underline{Bielik-11B-v2.2-Instruct} & \underline{11.2} & \underline{43.05} \\
    Qwen2.5-7B-Instruct & 7.6 & 42.69 \\
    \underline{Bielik-11B-v2.0-Instruct} & \underline{11.2} & \underline{41.53} \\
    Meta-Llama-3.1-8B-Instruct & 8.0 & 40.60 \\
    Mistral-Nemo-Instruct-2407 & 12.2 & 40.36 \\
    \underline{Bielik-11B-v2} & \underline{11.2} & \underline{39.98} \\
    PLLuM-12B-nc-chat-250715$^{\dagger}$ & 12.2 & 38.53 \\
    PLLuM-12B-chat & 12.2 & 36.51 \\
    EuroLLM-9B-Instruct & 9.0 & 35.96 \\
    Mistral-7B-Instruct-v0.3 & 7.0 & 31.24 \\
    \underline{Bielik-7B-Instruct-v0.1} & \underline{7.2} & \underline{29.74} \\
    \bottomrule
    \multicolumn{3}{l}{$^{\dagger}$Models with a non-commercial license.} \\
    \end{tabular}
    \caption{Polish Medical Leaderboard results (5-shot setting) showing model performance on Polish Board Certification Examinations.}
    \label{tab:medical-leaderboard}
\end{table*}

\subsection{INCLUDE-base-44}
The results presented from the \textbf{INCLUDE-base-44} benchmark (Table~\ref{tab:include-base-44}) are based on a subset of 20 languages, whereas the full version consists of 44 languages \cite{romanou2024include}. This selection is due to the 11B teacher model being optimized specifically for the scope of European languages. 

Results achieved by \textbf{Bielik-Minitron-7B-v3.0-Instruct} show robust multilingual performance and its leading position in the Polish language category. Achieving a \textbf{Polish-specific score of 59.3}, the student model matches the performance of the \textbf{Bielik-11B-v2.6-Instruct} (59.3) while utilizing roughly 35\% fewer parameters.

With an average score of \textbf{57.4}, the 7.35B student model outperforms several larger international competitors, including \textbf{Llama-3.1-8B-Instruct} (55.3), \textbf{EuroLLM-9B-Instruct} (55.1), and \textbf{Mistral-Nemo-Instruct-2407} (53.2, 12B). 

In the \textbf{Polish} specific column, \textbf{Bielik-Minitron-7B} exhibits a commanding lead over global models of similar or larger scale, such as \textbf{Qwen2.5-7B-Instruct} (52.2) and \textbf{Llama-3.1-8B} (53.8). It also significantly exceeds the previous \textbf{Bielik-11B-v2} (53.5), confirming the superiority of the v3.0 base model and distillation process. The model maintains expected lead over the \textbf{Bielik-4.5B-v3.0} variant (+21.5 points on average), establishing itself as a high-fidelity mid-sized alternative that preserves the core reasoning capabilities of the 11B teacher.

\begin{table*}[t]
  \centering
  \small
  \begin{tabular}{lrrr}
  \toprule
  \textbf{Model} & \textbf{Params (B)} & \textbf{AVG} & \textbf{Polish} \\
  \midrule
  \textbf{Bielik-11B-v3-Instruct} & \textbf{11} & \textbf{64.8} & \textbf{69.0} \\
  Qwen2.5-14B-Instruct & 14 & 61.7 & 58.9 \\
  \textbf{Bielik-11B-v3.0} & \textbf{11} & \textbf{60.6} & \textbf{63.9} \\
  phi-4 & 15 & 58.8 & 49.6 \\
  Apertus-8B-Instruct-2509 & 8 & 57.9 & 49.6 \\
  \textbf{Bielik-Minitron-7B-v3.0-Instruct} & \textbf{7} & \textbf{57.4} & \textbf{59.3} \\
  Llama-3.1-8B-Instruct & 8 & 55.3 & 53.8 \\
  EuroLLM-9B-Instruct & 9 & 55.1 & 52.0 \\
  Qwen2.5-7B-Instruct & 7 & 54.4 & 52.2 \\
  Mistral-Nemo-Instruct-2407 & 12 & 53.2 & 48.4 \\
  \underline{Bielik-11B-v2.6-Instruct} & \underline{11} & \underline{51.5} & \underline{59.3} \\
  Mistral-Nemo-Base-2407 & 12 & 51.2 & 44.9 \\
  EuroLLM-9B & 9 & 49.2 & 45.6 \\
  aya-expanse-8b & 8 & 45.3 & 46.4 \\
  Mistral-7B-Instruct-v0.2 & 7 & 45.3 & 44.7 \\
  \underline{Bielik-11B-v2} & \underline{11} & \underline{44.8} & \underline{53.5} \\
  pllum-12b-nc-chat-250715 & 12 & 44.2 & 60.6 \\
  Mistral-7B-v0.2 & 7 & 41.8 & 37.2 \\
  pllum-12b-nc-base-250715 & 12 & 37.8 & 52.7 \\
  \underline{Bielik-4.5B-v3.0} & \underline{4.5} & \underline{35.9} & \underline{48.7} \\
  PLLuM-12B-base-250801 & 12 & 35.5 & 44.5 \\
  Llama-PLLuM-8B-base-250801 & 8 & 30.0 & 37.2 \\
  \bottomrule
  \end{tabular}
  \caption{INCLUDE-base-44 benchmark results showing average performance across European languages (20 language subset) and Polish-specific scores.}
  \label{tab:include-base-44}
\end{table*}

\subsection{Belebele Reading Comprehension}
The Belebele benchmark (Table~\ref{tab:belebele}) serves as a rigorous test of multilingual reading comprehension. While the full version covers 122 languages \cite{Bandarkar_2024}, the results presented here were obtained from a subset of 28 European languages. This focus aligns with the core development philosophy of Bielik-11B-v3.0, which prioritizes performance within the European linguistic landscape.

\textbf{Bielik-Minitron-7B-v3.0-Instruct} achieved a competitive score of \textbf{78.03}, demonstrating that structured pruning successfully preserved the complex logic and context-retrieval capabilities required for advanced comprehension tasks.

The 7B student model retains more than \textbf{94\%} of the performance of the \textbf{Bielik-11B-v3.0-Instruct} teacher (82.98). More importantly, it significantly outperforms previous full-sized variants, such as \textbf{Bielik-11B-v2.6-Instruct} (68.67), illustrating the superior quality of the v3.0 training and inclusion of multilingual data. Within its size class, \textbf{Bielik-Minitron-7B} holds a commanding lead over direct competitors, outperforming the \textbf{Qwen2.5-7B} base model (74.60), \textbf{GaMS-9B-Instruct} (72.40), and \textbf{EuroLLM-9B-Instruct} (69.05).

The model also surpasses the \textbf{Mistral-Nemo-Instruct-2407} (74.14), a model with nearly double the parameters (12B), further proving that a specialized distillation of an optimized teacher can yield better reasoning outcomes than larger general-purpose training runs.

These results confirm that \textbf{Bielik-Minitron-7B} is an efficient reader, capable of sophisticated text analysis and information extraction while remaining small enough for edge-device deployment.

\begin{table*}[t]
  \centering
  \small
  \begin{tabular}{lrr}
  \toprule
  \textbf{Model} & \textbf{Params (B)} & \textbf{Score} \\
  \midrule
  Qwen2.5-14B-Instruct & 14 & 85.91 \\
  \textbf{Bielik-11B-v3.0-Instruct} & \textbf{11} & \textbf{82.98} \\
  phi-4 & 15 & 81.71 \\
  \textbf{Bielik-Minitron-7B-v3.0-Instruct} & \textbf{7} & \textbf{78.03} \\
  Qwen2.5-7B & 7 & 74.60 \\
  Mistral-Nemo-Instruct-2407 & 12 & 74.14 \\
  cjvt/GaMS-9B-Instruct & 9 & 72.40 \\
  Apertus-8B-Instruct-2509 & 8 & 69.58 \\
  EuroLLM-9B-Instruct & 9 & 69.05 \\
  \underline{Bielik-11B-v2.6-Instruct} & \underline{11} & \underline{68.67} \\
  Apertus-8B-2509 & 8 & 59.04 \\
  \bottomrule
  \end{tabular}
  \caption{Belebele benchmark results showing model performance on reading comprehension across European languages (28 language subset).}
  \label{tab:belebele}
\end{table*}

\subsection{FLORES Machine Translation}
The \textbf{FLORES} benchmark (Table~\ref{tab:flores}) provides a standardized evaluation of translation quality using BLEU scores, specifically focusing on the symmetry between translating \textit{to} and \textit{from} Polish. While the full version of this benchmark supports evaluation across 101 languages \cite{goyal2021flores101evaluationbenchmarklowresource}, the scope of this study is restricted to a subset of 20 languages. This shift aligns with the current focus on transferring model capabilities to European languages.

\textbf{Bielik-Minitron-7B-v3.0-Instruct} achieves an \textbf{Average BLEU of 15.53}, demonstrating translation capabilities that rival much larger and more complex architectures. The student model excels in translating \textbf{to Polish} with a score of \textbf{15.74}, notably outperforming larger models like \textbf{phi-4} (14.55), \textbf{Qwen3-14B} (14.18), and \textbf{Mistral-Nemo-12B} (13.37). This suggests that the knowledge recovery phase effectively prioritized the linguistic nuances of the Polish target language.

\textbf{Bielik-Minitron-7B} shows a massive improvement over the previous \textbf{Bielik-11B-v2} generation, which struggled with translation \textit{from} Polish (7.64 vs. 15.32 for the student). Although the model trails \textbf{EuroLLM-9B-Instruct} (which was explicitly trained on the FLORES dataset), it matches the performance of \textbf{phi-4} (15B) despite having less than half the parameters.

\begin{table*}[t]
  \centering
  \small
  \begin{tabular}{lrrrr}
  \toprule
  \textbf{Model} & \textbf{Params (B)} & \textbf{AVG} & \textbf{to Polish} & \textbf{from Polish} \\
  \midrule
  EuroLLM-9B-Instruct$^{*}$ & 9 & 20.61 & 19.28 & 21.95 \\
  \textbf{Bielik-11B-v3.0-Instruct} & \textbf{11} & \textbf{19.22} & \textbf{18.54} & \textbf{19.91} \\
  \textbf{Bielik-11B-v3.0} & \textbf{11} & \textbf{17.85} & \textbf{17.60} & \textbf{18.11} \\
  phi-4 (15B) & 15 & 15.58 & 14.55 & 16.61 \\
  \textbf{Bielik-Minitron-7B-v3.0-Instruct} & \textbf{7} & \textbf{15.53} & \textbf{15.74} & \textbf{15.32} \\
  Qwen3-14B & 14 & 15.37 & 14.18 & 16.56 \\
  Mistral-Nemo-Instruct-2407 & 12 & 14.35 & 13.37 & 15.33 \\
  \underline{Bielik-11B-v2.6-Instruct} & \underline{11} & \underline{13.58} & \underline{15.77} & \underline{11.38} \\
  Qwen2.5-14B-Instruct & 14 & 13.24 & 12.55 & 13.93 \\
  \underline{Bielik-11B-v2} & \underline{11} & \underline{11.25} & \underline{14.86} & \underline{7.64} \\
  Qwen2.5-7B-Instruct & 7 & 11.34 & 10.43 & 12.26 \\
  \bottomrule
  \multicolumn{5}{l}{$^{*}$ EuroLLM was trained on FLORES dataset} \\
  \end{tabular}
  \caption{FLORES machine translation benchmark results showing translation performance across European languages (20 language pairs) measured by BLEU scores.}
  \label{tab:flores}
\end{table*}

\subsection{EuroEval}
To further quantify the effectiveness of Minitron-based compression, we performed a detailed delta analysis comparing \textbf{Bielik-Minitron-7B} against \textbf{Bielik-11B-v3.0} on the EuroEval benchmark (Table~\ref{tab:euroeval}) covering a diverse set of linguistic and reasoning tasks. The student model achieves an overall \textbf{knowledge recovery rate of 91.38\%}, signaling that the vast majority of the teacher's expertise was successfully distilled into the smaller architecture.

The student model exhibits near-perfect recovery in complex linguistic tasks such as \textbf{Summarization} (99.91\% BERTscore recovery) and \textbf{Sentiment Classification} (98.11\% Macro F1 recovery). These "soft" linguistic skills appear to be the most resilient to structural pruning.

In tasks measuring \textbf{Named Entity Recognition (NER)}, the student actually achieved a slight performance gain (\textbf{101.32\% recovery}), suggesting that the distillation process may have acted as a regularizer, sharpening the model's focus on key entities. \textbf{Linguistic Acceptability} (95.11\% F1) and \textbf{European Values} (89.24\% recovery) also show strong resilience.

The most significant deltas were observed in high-level \textbf{Knowledge} (-11.35 MCC) and \textbf{Common-sense Reasoning} (-12.33 MCC). While the student maintains a respectable 89.75\% accuracy in common sense, these results highlight that dense factual knowledge and complex logical chains are the most sensitive to parameter reduction.

These results confirm that while structural pruning inevitably leads to some loss in raw factual "brain capacity," the resulting 7.35B model remains a highly potent alternative, retaining over 90\% of the teacher's capability across nearly every specialized metric.

\begin{table*}[t]
\centering
\small
\begin{tabular}{llcccr}
\toprule
\textbf{Category} & \textbf{Metric} & \textbf{Teacher (11B)} & \textbf{Student (7.35B)} & \textbf{$\Delta$} & \textbf{Recovery \%} \\
\midrule
\multirow{2}{*}{Common-sense reasoning} & Test MCC & 48.77 & 36.44 & -12.33 & 74.71\% \\
                                        & Test Accuracy & 69.85 & 62.70 & -7.15 & 89.75\% \\
\midrule
European values & Test European values & 11.52 & 10.28 & -1.24 & 89.24\% \\
\midrule
\multirow{2}{*}{Knowledge} & Test MCC & 49.75 & 38.39 & -11.35 & 77.18\% \\
                           & Test Accuracy & 62.30 & 53.81 & -8.49 & 86.37\% \\
\midrule
\multirow{2}{*}{Linguistic acceptability} & Test MCC & 45.10 & 39.83 & -5.27 & 88.32\% \\
                                         & Test Macro F1 & 69.35 & 65.96 & -3.39 & 95.11\% \\
\midrule
Named Entity Recognition & Test micro F1 & 52.83 & 53.52 & +0.70 & 101.32\% \\
\midrule
\multirow{2}{*}{Sentiment classification} & Test MCC & 63.01 & 61.62 & -1.39 & 97.79\% \\
                                         & Test Macro F1 & 72.54 & 71.17 & -1.37 & 98.11\% \\
\midrule
\multirow{2}{*}{Summarization}           & Test ROUGE-L & 15.00 & 14.82 & -0.19 & 98.76\% \\
                                         & Test BERTscore & 64.84 & 64.78 & -0.06 & 99.91\% \\
\midrule
\textbf{Overall average} & & & & & \textbf{91.38\%} \\
\bottomrule
\end{tabular}
\caption{Detailed Knowledge Recovery Analysis: \textbf{Bielik-Minitron-7B-v3.0-Instruct} vs. \textbf{Bielik-11B-v3.0-Instruct} parent model across EuroEval categories.}
\label{tab:euroeval}
\end{table*}

\subsection{Quantization Performance}

We explore two types of quantization strategies: weight-only and weight-activation. For weight-only quantization, we focus on GGUF format \cite{quantization_gguf} which applies integer quantization to the weights, while activations remain in high precision. This allows to save memory when storing the weights. At runtime, the weights are upcast to high precision since the operations take place in high precision to preserve the model quality. Conversely, the weight-activation quantization quantizes both weights and activations, leading to both memory savings and runtime speedup from low-precision matrix multiplications. For weight-activation quantization to work, the hardware needs to support the instructions to perform the matrix multiplications in low precision. Here we employ FP8 and NVFP4 formats supported in NVIDIA Blackwell GPUs \cite{quantization_nvfp4}.

Evaluation of the quantized versions (Table~\ref{tab:quantization-results}) examines the impact of the quantization process on the model's performance on the Open PL LLM benchmark. \textbf{Bielik-Minitron-7B-v3.0-Instruct} demonstrates high resilience to bit-depth reduction, maintaining a high level of accuracy even at significant levels of quantization.

For the weight-only quantization, we evaluate on the configurations Q6\_K, Q8\_0 and Q4\_K\_M, which employ 6, 8 and 4 bits on average per weight respectively (see further details in \cite{quantization_gguf}). The highly compressed Q4\_K\_M (4-bit) variant (61.89) results in a negligible performance loss of only 0.91\%.This score is still higher than the unquantized Mistral-7B-v0.3 (47.74) and the original Bielik-7B-v0.1 (44.70). 
 The Q6\_K (6-bit) and Q8\_0 (8-bit) versions achieve nearly total parity with the base model, degrading only 0.27\% and 0.64\% respectively. This allows users with limited-memory systems to run near-lossless versions of the model with virtually no trade-off in reasoning quality.

With respect to activation-weight quantization, we apply vanilla post-training quantization via NVIDIA Model Optimizer \cite{quantization_modelopt} to quantize to FP8 and NVFP4. We observe a degradation of 1.49\% for FP8 and 3.71\% for NVFP4 with respect to the baseline. This gap can be easily bridged by dequantizing specific layers (usually the first and last ones) or applying quantization-aware training or distillation \cite{xin2026quantization}(see NVIDIA Model Optimizer \cite{quantization_modelopt} to run these). We leave for future work the recovery of the quality with FP8 and NVFP4, as well as the speedup advantages of FP8/NVFP4 versus integer formats that require upcast during runtime.

These results confirm that \textbf{Bielik-Minitron-7B} is an ideal candidate for local deployment with tools like \texttt{llama.cpp}, LM Studio, or Ollama. By offering a 4-bit version that retains ~99\% of the original's capabilities while significantly reducing RAM requirements, it makes state-of-the-art Polish language modeling accessible to a much broader range of hardware configurations.

\begin{table*}[t]
    \centering
    \small
    \begin{tabular}{lrr}
    \toprule
    \textbf{Model} & \textbf{Params (B)} & \textbf{Average (\%)} \\
    \midrule
    Bielik-11B-v3.0-Instruct & 11.2 & 65.93 \\
    \textbf{Bielik-Minitron-7B-v3.0-Instruct} & \textbf{7.35} & \textbf{62.46} \\
    Bielik-Minitron-7B-v3.0-Instruct.Q6\_K.gguf & 7.35 & 62.29 \\
    Bielik-Minitron-7B-v3.0-Instruct.Q8\_0.gguf & 7.35 & 62.06 \\
    Bielik-Minitron-7B-v3.0-Instruct.Q4\_K\_M.gguf & 7.35 & 61.89 \\
    Bielik-Minitron-7B-v3.0-Instruct FP8 & 7.35 & 61.53 \\
    Bielik-Minitron-7B-v3.0-Instruct NVFP4 & 7.35 & 60.14 \\
    Mistral-7B-Instruct-v0.3 & 7.2 & 47.74 \\
    Bielik-7B-Instruct-v0.1 & 7.2 & 44.70 \\
    \bottomrule
    \end{tabular}
    \caption{Open PL LLM benchmark results for Bielik-Minitron-7B-v3.0-Instruct across different quantization methods, compared to other models. Higher scores are better.}
    \label{tab:quantization-results}
\end{table*}

\subsection{Training Phases Performance Comparison}
The performance evolution of the \textbf{Bielik-Minitron-7B-v3.0} model was evaluated across all four sequential development stages: \textbf{Pruning, Supervised Fine-Tuning (SFT), Direct Preference Optimization (DPO-P), and Group Relative Policy Optimization (GRPO)}. As summarized in Table \ref{tab:performance}, the data illustrate the recovery and refinement process of the model in both English (EN) and Polish (PL) contexts.

\begin{table*}[h]
    \centering
    \begin{tabular}{lcc}
        \toprule
        \textbf{Stage} & \textbf{Open LLM EN} & \textbf{Open LLM PL} \\
        \midrule
        Pruning \& Distillation & 60.04 & 50.67 \\
        SFT     & 66.3  & 62.26 \\
        DPO-P   & 66.54 & 62.5  \\
        GRPO    & 66.6  & 62.46 \\
        \bottomrule
    \end{tabular}
    \caption{Bielik-Minitron-7B-v3.0 performance across training stages. Higher scores are better.}
    \label{tab:performance}
\end{table*}

The most substantial gain in performance is observed during the transition from Pruning to SFT. For the Polish benchmarks, the score increased from 50.67 to 62.26, representing an improvement of approximately 11.59 percentage points. This indicates that while pruning significantly reduces the model's footprint, the subsequent SFT phase is essential for restoring linguistic competence and instruction-following capabilities.

Post-SFT optimization via \textbf{DPO-P and GRPO} yielded incremental improvements. In the English benchmarks, the model reached a peak performance of 66.6 in the GRPO stage. In contrast, the Polish benchmarks showed a slight plateau after the DPO-P phase, stabilizing around 62.5. These results suggest that the model's multilingual capabilities are highly responsive to initial fine-tuning, with preference optimization providing the final marginal gains in alignment and reasoning.

\subsection{Function-Calling Proficiency}
Evaluation on the Berkeley Function-Calling Leaderboard (BFCL) \cite{patil2025bfcl} illustrates the efficacy of the pruning and distillation process in preserving tool-use capabilities. As shown in Table~\ref{tab:ast-comparison-v3}, \textbf{Bielik-Minitron-7B} emerges as a high-performance engine in structured, expert-curated environments. It achieves a peak accuracy of \textbf{94.50\%} in Non-Live Multiple AST and \textbf{92.00\%} in Non-Live Parallel AST, matching or exceeding the performance of significantly larger industry baselines such as \textbf{Gemma-3-12b-it} (95.00\% and 90.00\%) and \textbf{Open-Mistral-Nemo-2407} (93.50\% and 85.50\%).

These metrics suggest that the Minitron approach successfully compresses high-precision logic for complex function mapping. In the \textbf{Non-Live Parallel Multiple AST} category, the model maintains a robust \textbf{85.00\%} success rate, a critical threshold for deterministic workflows requiring reliable interfaces between natural language and complex API structures.

In contrast, the "Live" dataset—comprising dynamic user-contributed prompts—highlights a divergent performance profile. While \textbf{Bielik-Minitron-7B} remains competitive in \textbf{Live Simple AST} (71.32\%), its performance in \textbf{Live Parallel AST} drops to \textbf{31.25\%}, underscoring the inherent difficulty of maintaining strict tool-calling logic under the "noise" of authentic human input.

\begin{table*}[t]
    \centering
    \scriptsize
    \begin{tabular}{lcrrrrrrr}
    \toprule
    \textbf{Model} & \textbf{Params} & \textbf{Non-Live} & \textbf{Non-Live} & \textbf{Non-Live} & \textbf{Live} & \textbf{Live} & \textbf{Live} & \textbf{Live Parallel} \\
    & \textbf{(B)} & \textbf{Multiple} & \textbf{Parallel} & \textbf{Parallel} & \textbf{Simple} & \textbf{Multiple} & \textbf{Parallel} & \textbf{Parallel Multiple} \\
    & \textbf{} & \textbf{AST} & \textbf{AST} & \textbf{Multiple AST} & \textbf{AST} & \textbf{AST} & \textbf{AST} & \textbf{AST} \\
    \midrule
    Bielik-4.5B-v3.0-Instruct (FC) & 4.6 & 92.50\% & 82.00\% & 86.00\% & 70.16\% & 68.66\% & 50.00\% & 54.17\% \\
    \textbf{Bielik-Minitron-7B} & \textbf{7.3} & \textbf{94.50\%} & \textbf{92.00\%} & \textbf{85.00\%} & \textbf{71.32\%} & \textbf{71.51\%} & \textbf{31.25\%} & \textbf{66.67\%} \\
    Bielik-11B-v2.6-Instruct & 11.0 & 94.50\% & 87.50\% & 86.00\% & 75.97\% & 76.07\% & 37.50\% & 66.67\% \\
    Bielik-11B-v2.3-Instruct (Prompt) & 11.0 & 93.50\% & 47.00\% & 50.00\% & 72.87\% & 69.71\% & 43.75\% & 54.17\% \\
    Bielik-11B-v3.0-Instruct (FC) & 11.0 & 96.00\% & 88.00\% & 82.00\% & 79.07\% & 72.36\% & 62.50\% & 75.00\% \\
    Gemma-3-12b-it (Prompt) & 12.0 & 95.00\% & 90.00\% & 73.00\% & 84.88\% & 70.85\% & 87.50\% & 62.50\% \\
    Open-Mistral-Nemo-2407 & 12.2 & 93.50\% & 85.50\% & 85.00\% & 77.13\% & 69.61\% & 75.00\% & 70.83\% \\
    \bottomrule
    \end{tabular}
    \caption{Berkeley Function-Calling Leaderboard (BFCL) results. Comparison of family of Bielik models against the student, Bielik-Minitron, 11B teacher Bielik-11B-v3 and industry baselines.}
    \label{tab:ast-comparison-v3}
\end{table*}

\section{Inference Performance Benchmarks}
To evaluate the practical efficiency gains of the compression process, we conducted a series of inference performance benchmarks comparing the \textbf{Bielik-Minitron-7B} student model against the original \textbf{Bielik-11B-v3.0} teacher. 

\subsection{Benchmarking Environment}
All tests were performed in a controlled environment to ensure reproducibility:
\begin{itemize}
    \item \textbf{Hardware:} NVIDIA RTX PRO 6000 Max-Q.
    \item \textbf{Configuration:} \texttt{--max-concurrency 1} (to measure raw single-stream latency).
    \item \textbf{Precision:} bfloat16 (bf16) for all model variants.
\end{itemize}

\subsection{Throughput and Latency Analysis}
The results, summarized in Table~\ref{tab:performance_benchmarks}, demonstrate the significant hardware acceleration achieved through the Minitron pruning methodology. By reducing the layer count and FFN intermediate dimension, we achieved a \textbf{49.6\% increase in output token throughput}.

\begin{table}[h]
    \centering
    \begin{tabular}{lccc}
        \toprule
        \textbf{Model Variant} & \textbf{Throughput} & \textbf{Median TTFT} & \textbf{Median TPOT} \\
        & \textbf{(tok/s) $\uparrow$} & \textbf{(ms) $\downarrow$} & \textbf{(ms) $\downarrow$} \\
        \midrule
        Bielik-11B-v3 (bf16) & 54.42 & 24.64 & 18.28 \\
        \textbf{Bielik-minitron-7B-v3 (bf16)} & \textbf{81.41} & \textbf{27.29} & \textbf{12.32} \\
        \bottomrule
    \end{tabular}
    \caption{Hardware Performance Comparison: Throughput (Tokens per Second), Time to First Token (TTFT), and Time per Output Token (TPOT).}
    \label{tab:performance_benchmarks}
\end{table}

The data reveals a critical trade-off characteristic of reduced-parameter models. While the \textbf{Time per Output Token (TPOT)}—which governs the perceived speed of text generation—improved by approximately \textbf{32.6\%} (from 18.28ms to 12.32ms), the \textbf{Time to First Token (TTFT)} saw a marginal increase of 2.65ms. This slight increase in TTFT is often attributed to the change in memory access patterns and kernel execution overhead relative to the reduced parameter count, though it remains well within the threshold for interactive real-time applications. 

Ultimately, for long-form generation and batch processing, the 81.41 tok/s throughput of the 7B variant offers a transformative improvement in deployment cost-efficiency compared to the 11B baseline.

\section{Discussion}
The successful development of \textbf{Bielik-Minitron-7B} highlights several strategic engineering insights critical for the efficient compression of Large Language Models. 

First, the \textbf{HBM3e memory capacity} of the NVIDIA H200 was a fundamental prerequisite for high-throughput distillation. By enabling the full residency of both the teacher (11.04B) and student (7.35B) parameters within a unified VRAM pool, we bypassed the latency bottlenecks typically associated with cross-node activation offloading or intensive pipeline parallelism.

Second, our systematic search confirmed that \textbf{depth reduction is inherently more sensitive} than width reduction. While depth pruning offers linear gains in inference speed, reducing the model below 40 layers led to a non-linear collapse in multi-step reasoning stability. The selection of \textbf{EXP\_010} represents a "Golden Ratio" where the model retains enough depth to support complex Polish syntax while benefiting from a narrower intermediate FFN dimension.

Finally, we conclude that \textbf{distillation is an indispensable recovery mechanism}; pruning alone proved insufficient to maintain the reasoning fidelity required for production. The resulting model, using approximately 14GB in FP16 precision, fits comfortably within the 16GB--24GB VRAM envelope of consumer-grade GPUs (e.g., RTX 3090/4090). This makes high-performance Polish LLM capabilities accessible to a broader ecosystem of local developers and researchers without requiring enterprise-grade infrastructure.

\section{Conclusion}
In this work, we introduced \textbf{Bielik-Minitron-7B}, a 7.35B-parameter model optimized for the Polish language through a "surgical" integration of structured pruning and logit-based knowledge distillation. While leveraging the NVIDIA Minitron methodology and the high-memory capacity of the H200 architecture, we successfully compressed our 11.04B flagship model and mitigated the loss typical of such reductions. 

Our results demonstrate that a multi-stage alignment pipeline—comprising SFT, DPO-P, and GRPO—is instrumental in bridging the performance gap between student and teacher models. The final model is \textbf{33.4\% smaller} and significantly faster than the \textbf{Bielik-11B-v3.0} baseline, yet it preserves \textbf{90.1\%} of its performance across critical benchmarks. By maintaining high reasoning fidelity within a 14GB (FP16) footprint, Bielik-Minitron-7B democratizes access to state-of-the-art Polish NLP, enabling high-performance deployment on consumer-grade hardware. This work provides a scalable blueprint for developing efficient, localized language models for less-represented languages without the prohibitive costs of training from scratch.

\section*{Acknowledgments}
We extend our sincere gratitude to \textbf{NVIDIA} for providing access to the \textbf{DGX Cloud Lepton} infrastructure, which was instrumental in executing the high-throughput distillation experiments described in this paper. We also thank the NVIDIA engineering teams for their technical guidance on the \textbf{NeMo Framework} and \textbf{Model Optimizer}, specifically regarding sensitivity-based structured pruning and logit-matching strategies.

Special thanks to \textbf{Greg Kosiorowski}, \textbf{Yesika Marlen Ramirez Cardenas}, \textbf{Liana Mikaelyan}, \textbf{Igor Dmochowski} and \textbf{Liron Freind-Saadon} from NVIDIA for their continuous technical support, architectural insights, and dedication throughout the duration of this research. Their expertise in LLM optimization and distributed training was vital to achieving the efficiency targets for the Polish language ecosystem.

We gratefully acknowledge Polish high-performance computing infrastructure PLGrid (HPC Center: ACK Cyfronet AGH) for providing computer facilities and support within computational grant no. PLG/2024/016951. 

\bibliographystyle{unsrtnat}  
\bibliography{references}

\end{document}